\documentclass[10pt,twocolumn,letterpaper]{article}

\usepackage{cvpr}
\usepackage{times}
\usepackage{epsfig}
\usepackage{graphicx}
\usepackage{amsmath}
\usepackage{amssymb}
\usepackage{subfig}

\cvprfinalcopy 

\def\cvprPaperID{****} 

\ifcvprfinal\fi
\begin{document}

\title{Unsupervised learning of depth and motion}

\author{Kishore Konda\\
Goethe University Frankfurt\\
Germany\\
{\tt\small konda@informatik.uni-frankfurt.de}
\and
Roland Memisevic\\
University of Montreal\\
Canada\\
{\tt\small roland.memisevic@umontreal.ca}
}

\maketitle

\begin{abstract}
We present a model for the joint estimation of disparity and motion.  The model is based on learning about the interrelations between images from multiple cameras, multiple frames in a video, or the combination of both.  We show that learning depth and motion cues, as well as their combinations, from data is possible within a single type of architecture and a single type of learning algorithm, by using biologically inspired ``complex cell'' like units,  which encode correlations between the pixels across image pairs.  Our experimental results show that the learning of depth and motion makes it possible to achieve state-of-the-art performance in 3-D activity analysis, and to outperform existing hand-engineered 3-D motion features by a very large margin. 
\end{abstract}

\section{Introduction}
\label{intro}
A common property of 3-D inference and motion estimation is that both 
rely on establishing correspondences between pixels in two (or more) images. 
For depth estimation, these are correspondences between multiple views of a scene, 
for motion estimation between multiple frames in a video. 
Despite superficial differences between these tasks, such as 
the typical size of the average displacement across images, 
or whether the geometry is constant or variable across pairs, there are much stronger 
commonalities between the tasks, such as the fact that both rely on finding positions 
in one image which match those in another image. 
This suggests that both tasks may be learnable using essentially the same type 
of architecture and the same type of learning algorithm, 
but there has been hardly any work on trying to exploit this in practice.  
Besides the obvious advantage of allowing us to develop and maintain a single 
piece of code to achieve both tasks, 
it makes it trivial to fuse the information from both sources 
and thereby to design architectures that learn representations of multi-camera video 
streams with application, for example, in activity analysis. 


In the neuroscience literature, the so-called complex cell ``energy model'', 
is assumed to be the main underlying mechanism behind both depth and motion 
estimation (e.g. \cite{Adelson85,FleetBinocular}), 
and it provides an elegant explanation for how the brain can learn both using the same type 
of neural hardware.

There has been some progress recently in learning motion energy models from 
data \cite{ISA, Taylor:2010}, and learning based methods are among the state-of-the-art
in activity analysis from videos.  
However, there has been hardly any work on learning energy models for depth inference, 
nor for learning depth and motion information at the same time. 
In this work we show that it is, in fact, possible to \emph{learn} about 3-D depth 
entirely unsupervised from data, similar to learning motion as done using complex 
cell type models. 
Our experiments show how this makes it possible to 
achieve state-of-the-art performance in 3-D activity analysis from multi-camera video 
without making use of any hand-crafted features.

\subsection{Biologically inspired models of correspondence}
The first step to infer depth from two views is 
to find correspondences between points which represent the same 3-D location \cite{Hartley2004}. 
The two standard ways to approach this task are:  
1.) For each 
position in one image find a nearby matching point in the other image using some
measure of similarity between local image patches (e.g. \cite{scharstein2002taxonomy}). 
2.) For each position in both images, extract features that describe phase and 
frequency content of the region around that point, and read
off the phase difference across the two images from the set of filter 
responses \cite{qian1994computing}. 

The first approach has been more common in practice, although the second is more 
biologically plausible, as it does not require loops over local patches. 
More importantly, the second approach is amenable to \emph{data-driven learning} as 
we shall show. 
The most well-known account of phase-based disparity
estimation is the binocular energy or cross-correlation model (e.g. \cite{qian1994computing, fleet1991phase, ohzawa1990stereoscopic}).
In its most basic form, this model states that local disparities are encoded in the sum of the squared responses of
two neurons, each of which has a binocular receptive field. Each binocular receptive field, in turn,
shows a position-shift across the two views. Between them the two receptive fields show a quadrature
relationship (within each view). It can be shown that the position-shift across the views allows the
energy model to encode local disparity, while the quadrature relationship within each view allows it
to be independent of the Fourier phase of the local stimulus \cite{qian1994computing,FleetBinocular}.

Analogous models, also based on energy or cross-correlation, have been proposed independently 
for motion encoding \cite{FleetBinocular,OlshausesCadieu,multiview}. 
This is not surprising if one considers that motion can be defined 
as the transformation of a given input over time and disparity as the transformation of the input
across multiple views or a stereo pair. 
If the given input is a set of frames from a time sequence the model encodes motion
and when the input is a stereo pair it encodes disparity. It has been proposed 
that, due to similarity of models for motion and disparity encoding, it should be possible 
to integrate them \cite{qian1994computing}. But to date, there has been no practical 
exploration of this idea, nor of the \emph{learning} of depth from data. 

In this paper we present an approach to learning depth, motion and their combination from 
data, by using a feature learning architecture based on the energy model. 
Our approach is based on the view of energy model proposed by \cite{synchrony}, which 
shows that the (motion) energy model can be viewed as two independent contributions 
to motion encoding: 1) the detection of spatio-temporal ``synchrony'',  
and 2) the encoding of invariance. 
\cite{synchrony} present an autoencoder model using multiplicative interactions
for detection of synchrony, and they show that a pooling layer independently trained 
on the hidden responses can be used to achieve content invariance.
We adopt that approach for the estimation of depth and motion cues, 
as it gives rise to an efficient single-layer learning algorithm. But there a variety of learning 
based energy models that one could use instead (e.g., \cite{ISA, Taylor:2010}).

A description of the synchrony condition and how it can be used for implicit encoding of depth 
is presented in the next section. 
Since depth is encoded implicitly in the feature responses of the model, 
we then show how it is possible to ``calibrate'' an energy model learned on stereo data 
using available ground truth data to compute an explicit depth map from this encoding.  
Since in most applications, the representation of depth is a means to an end not a goal on 
its own, we then explore a variety of ways to utilize the implicit encoding of depth, as well as 
motion, using the same approach to learning features. 
We evaluate and compare several variations of this approach on the Hollywood3D activity recognition 
dataset \cite{Hadfield13}, and we demonstrate that it improves significantly upon 
the state-of-the-art, using a minimum of hand engineering. 

\section{Depth as a latent variable}
\label{depth}
The classic energy model (e.g. \cite{Adelson85,FleetBinocular}) states that we can obtain an estimate 
of the transformation, $P$, between two images $\vec{x}_1$ and $\vec{x}_2$ 
by computing a weighted sum over products of filter responses on the images. 
In particular, 
if the filters themselves differ by the transformation $P$, so that, 
\begin{equation}
\label{eq:synchrony}
\vec{w}_2 = P\vec{w}_1
\end{equation}
then the product filter responses will be large for input images for which $\vec{x}_2 = P\vec{x}_1$ holds, too. 
This makes it possible to extract motion, if $\vec{x}_1$ and $P\vec{x}_2$ denote adjacent frames 
in a video, and disparity if they denote two patches cropped from the same position 
of a stereo pair. 
In most practical situations (for both motion and disparity estimation),  
the dominant transformation between the images is a \emph{local translation}, 
in which case the optimal filters are Gabor features and $P$ is a small phase shift. 
In early, biologically motivated approaches to estimating displacements, filters have been hand-coded 
\cite{Adelson85,FleetBinocular}. In the context of motion estimation, various approaches 
were proposed recently to learning the filters from data (e.g. \cite{Taylor:2010,ISA,gated}). 
While learning has been inefficient due to the vast amounts of image patch pairs required for learning 
good filters, \cite{synchrony} recently presented the ``synchrony autoencoder'', which learns 
motion representations more efficiently, using a single-layer autoencoder with multiplicative 
interactions. 
We use a similar approach for defining models that learn to encode depth. 
We shall review that model, as well as show how we can use it for depth and motion estimation in the 
following section.

\subsection{Depth across stereo image pairs}
\label{stereo_pair_model}
Based on the above description, we can define a model based on the synchrony autoencoder (SAE) \cite{synchrony}
for learning depth representation from stereo pair of images as follows.
Assume we are given a set of stereo image pairs, $\vec{x}, \vec{y} \in\mathbb{R}^{N}$.  
Let $\mathbf{W}^x, \mathbf{W}^y \in\mathbb{R}^{Q\times N}$ denote the matrices containing 
$Q$ feature vectors $\vec{W}^x_q,\vec{W}^y_q \in\mathbb{R}^{N}$, stacked row-wise.

We define the latent representations $\vec{f}^x = \mathbf{W}^x\vec{x}$ and $\vec{f}^y = \mathbf{W}^y\vec{y}$,
which are typically called {\it factors} in the context of energy models \cite{Taylor:2010,gated,synchrony}.
The hidden representation of disparity is then defined as
\begin{equation}
\label{2framesae}
  \vec{h} = \sigma(\vec{f}^x\odot \vec{f}^y)
\end{equation}
where $\sigma=(1+\exp(-x))^{-1}$ is a saturating non-linearity. (We use the logistic sigmoid in this 
work, but other non-linearities could be used as well.)

\begin{figure}
 \centering
 \includegraphics[scale=0.45]{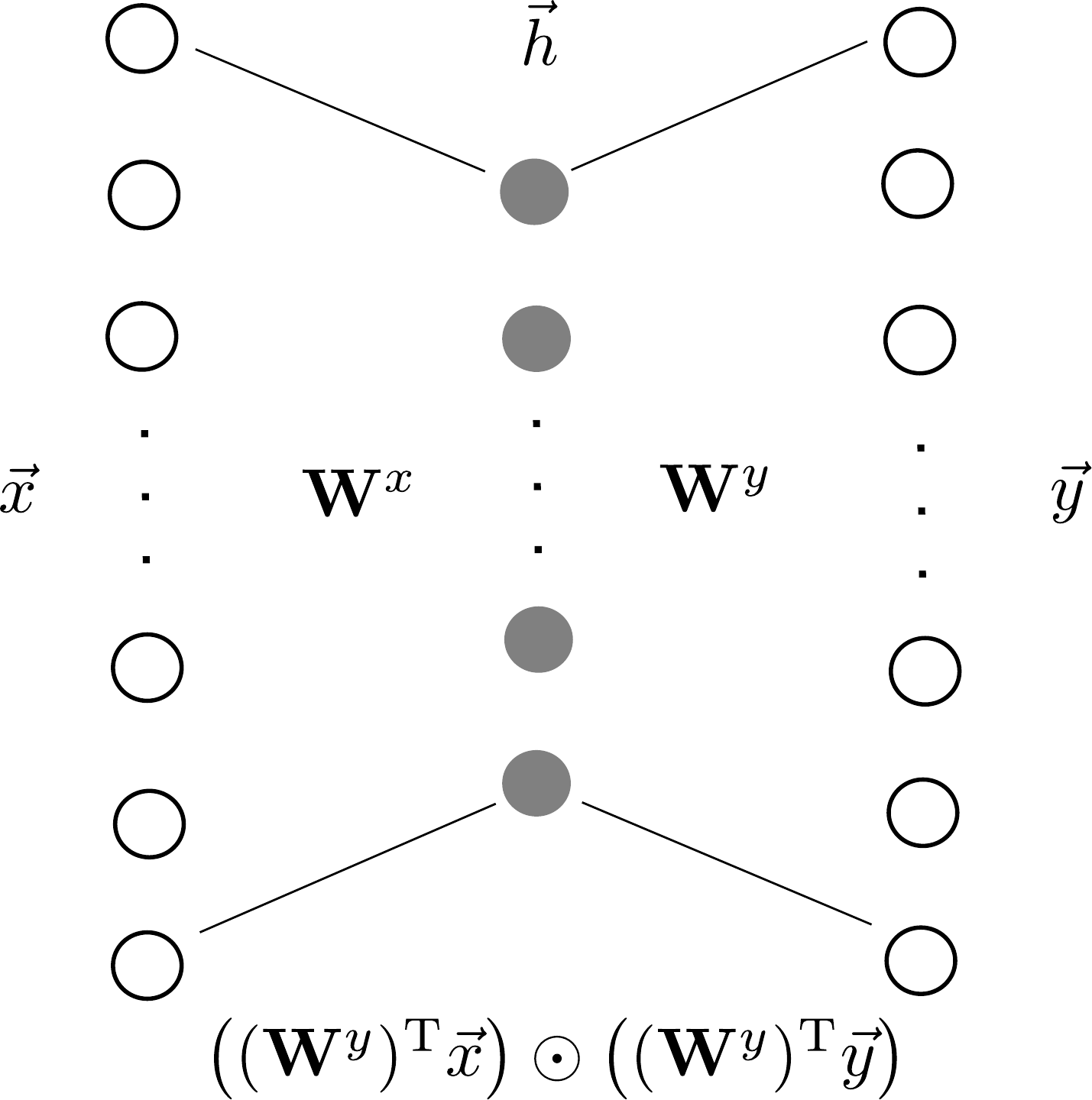}
 \caption{A model encoding the transformation inherent in the image pair ($\vec{x}, \vec{y}$).}
 \label{figure:sae_model}
\end{figure}

A standard way to train an autoencoder is by minimizing reconstruction error. 
Since the vector of multiplicative interactions between factors represents the transformation 
between $x$ and $y$, here we may define the reconstruction of one input given the 
other input \emph{and} the transformation as 
\begin{eqnarray}
\label{decoder}
\hat{x}= (\mathbf{W}^x)^\mathrm{T}(\vec{h} \odot \vec{f}^y) \\
\hat{y}= (\mathbf{W}^y)^\mathrm{T}(\vec{h} \odot \vec{f}^x)
\end{eqnarray}
Here, we assume an autoencoder with tied weights similar to \cite{Taylor:2010,gated,synchrony}.
This allows us to define the reconstruction error the as symmetric squared difference between inputs and 
their corresponding reconstructions: 
\begin{equation}
\label{reconstruction_error}
 L((\vec{x},\vec{y}),(\hat{\vec{x}},\hat{\vec{y}})) = \|(\vec{x} - \hat{\vec{x}})\|^2 + \|(\vec{y} - \hat{\vec{y}})\|^2
\end{equation}

\subsubsection{Regularization}
For extraction of sparse and robust representation we use contraction as regularization \cite{contractiveAE} which amounts to adding the 
Frobenius norm of the Jacobian of the hiddens with respect to the inputs $x,y$.
\begin{equation}
 \|J_e(\vec{x},\vec{y})\|^2_E = \sum_{ij}\left(\frac{\partial h_j(\vec{x},\vec{y})}{\partial x_i}\right)^2 + \sum_{ij}\left(\frac{\partial h_j(\vec{x},\vec{y})}{\partial y_i}\right)^2
\end{equation}
Using sigmoid non-linearity the contraction term becomes 
\begin{equation}
\begin{split}
\label{equation:contraction}
 \|J_e(\vec{x},\vec{y})\|^2_E  = &\sum_{j}(h_j(1-h_j))^2 (f^x_j)^2 \sum_{i}({W^x_{ij}})^2 \\
		  & + \sum_{j}(h_j(1-h_j))^2 (f^y_j)^2 \sum_{i}({W^y_{ij}})^2
\end{split}
\end{equation}

Thus the complete objective function employing contractive regularization, using $\lambda$ as the regularization strength, is
\begin{eqnarray}
\label{regularizedcost}
\mathcal{J}_{C}  = L((x,y),(\hat{x},\hat{y})) + \lambda \|J_e(x,y)\|^2_E
\end{eqnarray}

To obtain filters that represent depth we minimize Eq.~\ref{regularizedcost} for a set of image pairs cropped 
from identical positions of multiple views of the same scene. 
It is important to use a patchsize that is large enough to cover the maximal disparity in the data, 
otherwise the model will not be able to encode the corresponding depth. In contrast to traditional 
approaches to estimating depth, however, there is no need for rectification, since the model can learn 
any transformation between the frames not just horizontal shift.

\subsection{Depth across stereo sequences}
In the previous section we described a model for encoding depth across stereo image pairs.
We now propose several extensions of this approach to learn representations from stereo 
sequences not still images. This makes it possible to extract a representation informed 
by both motion and depth from the sequence. We defer the detailed quantitative evaluation 
of the approaches to Section~\ref{experiments}.

\subsubsection{Encoding depth}
Let $\vec{X},\vec{Y} \in\mathbb{R}^{N}$ be the concatenation of $T$ 
vectorized frames $\vec{x}_t,\vec{y}_t \in\mathbb{R}^{M}, t=1,\dots,T$,  
and be defined such that $(\vec{x}_t,\vec{y}_t)$ are stereo image pairs. 
Let $\mathbf{W}^x, \mathbf{W}^y \in\mathbb{R}^{Q\times N}$ denote matrices containing 
$Q$ feature vector pairs $\vec{W}_q^x, \vec{W}_q^y \in\mathbb{R}^{N}$ stacked row-wise.  
Each feature is composed of individual frame features $\vec{w}_{qt}^x\in\mathbb{R}^{M}$ 
each of which spans one frame $\vec{x}_{t}$ from the input sequence. 
Accordingly for the features in $\mathbf{W}^y$. 

In analogy to the previous section, we can define the factors  
$\vec{F}^X = \mathbf{W}^x\vec{X}$ and $\vec{F}^Y = \mathbf{W}^y\vec{Y}$ 
corresponding to the sequences $\vec{X}, \vec{Y}$.
A simple representation of depth 
may then be defined as
\begin{equation}
\label{depth_eq}
 H^D_q = \sigma(F^x_q \cdot F^y_q)
\end{equation}
The representation $H^D_q$ will contain products of frame responses 
$\big((\vec{w}_{qt}^x)^T\vec{x}_t\big) \cdot \big((\vec{w}_{qt}^y)^T\vec{y}_t\big)$ 
which detect synchrony over stereo pairs encoding depth. It will also contain 
products across time and position, 
$\big((\vec{w}_{qt}^x)^T\vec{x}_t\big) \cdot \big((\vec{w}_{q(t+i)}^y)^T\vec{y}_{t+i}\big)$,  
which will weakly encode motion as well. 
In other words, motion is encoded indirectly by this model, by computing products of responses 
at different times across cameras. 
We shall refer to this model as {\bf SAE-D} for ``depth encoding synchrony autoencoder'' in the following.


\subsubsection{Encoding motion}
For analyzing the effect of encoding motion vs. depth on the classification of sequences,  
we can define a hidden representation which employs only a single stereo sequence as follows. 
Let $\vec{X}$=$\vec{Y}$ represent a single camera channel from the stereo sequence. 
If we tie the weight matrices $\mathbf{W}^x, \mathbf{W}^y$ to be identical as well, 
Eq.~\ref{depth_eq} may be rewritten 
\begin{equation}
\label{motion_eq}
H^M_q = \sigma((F^x_q)^2)
\end{equation}
Since $F^x_q$ is the sum over individual frame filter responses, its square, by the 
binomial identity, will contain products of individual frame responses across 
time $\big((\vec{w}_{qt}^x)^T\vec{x}_t\big) \cdot \big((\vec{w}_{q(t+i)}^x)^T\vec{x}_{t+i}\big)$ 
as well as the squares of filter responses on individual frames. 
$H^M_q$ will therefore take on a large value only for those filters which match all 
indvidual frames, which implies that they will jointly satisfy Eq.~\ref{eq:synchrony}. 
This observation is the basis for the well-known equivalence between the energy model 
and the cross-correlation model (see, for example, \cite{FleetBinocular,multiview,synchrony}). 

Thus, in this case synchrony is detected over time, encoding the motion present in 
the input sequence. 
In analogy to the previous section, the encoding of motion will be weakly related 
to depth in the scene, as well, because depth and motion tend to be correlated. 
Any camera motion, for example, may be viewed as providing multiple views of 
a single scene, thereby implicitly containing information about depth (a fact that 
is exploited in structure-from-motion approaches). 
However, due to the absence of camera motion which is \emph{consistent} across the dataset, 
as well as the presence of a multitude of object motions, 
the depth information will only be weakly present in any encoding of motion. 
We shall call this model for representing motion {\bf SAE-M} in the following. 
The model is equivalent to the SAE defined in \cite{synchrony} for encoding 
motion from a single-channel video.

\subsubsection{Multiview disparity}
To obtain an explicit encoding of both depth and motion, we require the detection of 
synchrony both across time and across stereo-pairs.
One way to obtain such a representation in practice is to combine the representations 
defined in the previous two sections, for example, by using their average or concatenation. 

As a third alternative, we propose defining a joint representation by including products of 
frame responses across both time and stereo-pairs. 
Recall that the square of the sum over frame-wise filter responses contains 
within-channel motion information. We suggest obtaining an estimate of the across-channel 
disparity information by defining the hidden unit response as the product over 
theses squares. This allows us to extract information about disparity from the 
relation between the temporal evolutions of the complete video sequence, rather than 
between feature positions across single frames. 
To this end, we define the hidden representation 
\begin{equation}
\label{motion_depth_eq}
H^{MD}_q = \sigma((F^x_q)^2 \cdot (F^y_q)^2)
\end{equation}
The representation $H^{MD}_q$ may be written 
as $\big(\sum(\vec{w}_{qt}^x)^T\vec{x}_t\big)^2 \cdot \big(\sum(\vec{w}_{qt}^y)^T\vec{y}_t\big)^2$, 
and it may be thought of as a ``multi-view'' or ``motion-based'' estimate of disparity. 
We call a model based on this representation of disparity 
{\bf SAE-MD} in the following.

\subsection{Learning}
For the models described above the decoder and reconstruction cost can be derived to be 
similar to that of stereo-pair model in Section \ref{stereo_pair_model}. 
In particular, the reconstruction error and contraction 
cost for the models SAE-D and SAE-M can be derived by replacing the corresponding 
parameters of Equations~\ref{reconstruction_error} and \ref{equation:contraction}. 
For the SAE-D model this amounts to replacing frames $\vec{x},\vec{y}$ with sequences
$\vec{X},\vec{Y}$, and for the SAE-M model to further substituting $\vec{X}$ for $\vec{Y}$.

For the SAE-MD model, we found the contraction cost to be unstable due to presence of 
higher exponents in the hidden representation. Because of this, we use the trained weights 
from the SAE-D model and during inference use the representation 
from Equation \ref{motion_depth_eq}.
Alternatively it may be possible to train the model using a denoising criterion 
instead of contraction for regularization. 

\begin{figure}
 \centering
 \subfloat{\includegraphics[scale=1.08]{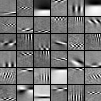}} \hspace{5mm}
 \subfloat{\includegraphics[scale=1.08]{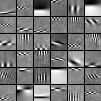}}
 \caption{Filters learned on stereo patch pairs from the KITTI dataset.}
 \label{stereo_pairs}
\end{figure}

\subsection{Interest point detector}
\label{IPD}
Hand-crafted image, motion or 4-D descriptors are typically accompanied by corresponding 
interest point detectors. Since they reduce the number of positions to extract representations 
from, they have been shown to improve efficiency and performance, for example, in bag-of-features 
based recognition pipelines. 

For a learned representation that is based on the linear projection of image patches, 
it is possible to define a default interest point operator, by 
using norm-thresholding of feature activations (see, for example, \cite{ISA}). 
It can be motivated by the observation that norms of relevant features will 
be higher at edge and motion locations than at homogeneous or static locations \cite{5539947}.
Norm thresholding interest point detection amounts to simply discarding features $H$ 
with norm $|H|_1 < \delta$. The value of $\delta$ may be chosen based on the
mean norm of the features in the training set.

\section{Experiments}
\label{experiments}
\subsection{Learning depth from image pairs}
It has been well-known that energy and cross-correlation models 
with hand-crafted Gabor features are able to extract depth information from 
random-dot stereograms (e.g., \cite{qian1994computing}).
In order to test whether depth information can be extracted in more realistic 
settings and using features that are learned from data, as proposed in Section~\ref{depth}, 
we first conducted an experiment where a depth map is estimated given a stereo image pair. 
For this experiment we use stereo images from the KITTI stereo/flow benchmark \cite{Geiger2012CVPR}. The dataset consists of 194 training 
image pairs and 195 test image pairs. 
For the training image pairs corresponding ground truth depth is provided. 
Since the ground truth is captured by means of a Velodyne sensor which is calibrated 
with the stereo pair it is only provided for approximately $30 \%$ of their
image pixels. We down-sampled the images from a 
resolution of $1226 \times 370$ pixels to $300 \times 100$ pixels, 
so that the local shift between image pairs falls within the local patch size, 
which is a crucial requirement for models using local phase matching 
for disparity computation as discussed in Section~\ref{depth}. 

We trained the stereo-pair model described 
in Section \ref{stereo_pair_model} (Eq.~\ref{2framesae}) 
on patch pairs cropped from the training set.  
Each patch is of size $16 \times 16$ pixels and
the total number of training samples is $100,000$.
The patches used for learning the filters are cropped only from regions of images 
where corresponding depth information is available. 

Some learned filters are shown in Figure \ref{stereo_pairs}. The figure shows that filters 
are localized, Gabor-like and span a wide range of frequencies and positions. Since 
cameras are parallel the filters learned predominantly horizontal shifts. 

To test if we can extract depth information from the learned hidden representation,  
we trained a logistic regression classifier 
using the available ground truth as the output data. 
To this end, we generate labels by taking the mean over non-zero pixel intensities 
of corresponding patches from the ground truth, which we then quantize into $25$ bins.  
After training the classifier, estimation of depth for a given stereo pair involves dense 
sampling of patch pairs followed by feature computation and prediction by the classifier. 

\begin{figure}
 \centering
 \subfloat[Left image]{\includegraphics[scale=0.5]{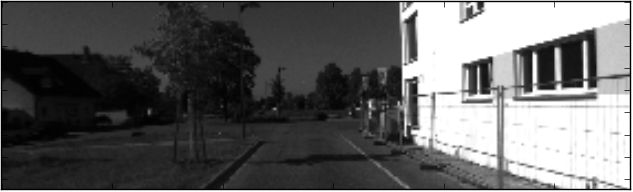}} \\
 \subfloat[Right image]{\includegraphics[scale=0.5]{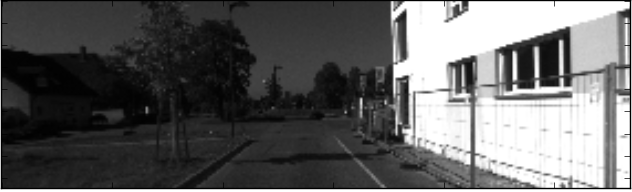}} \\
 \subfloat[Depth Map]{\label{masked}\includegraphics[scale=0.51]{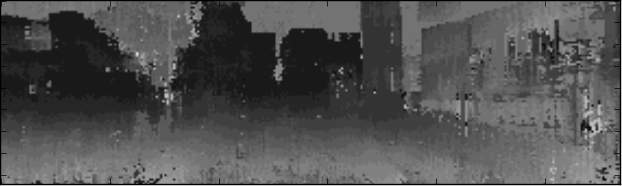}}\\
 \subfloat[Depth map masked using interest points]{\includegraphics[scale=0.51]{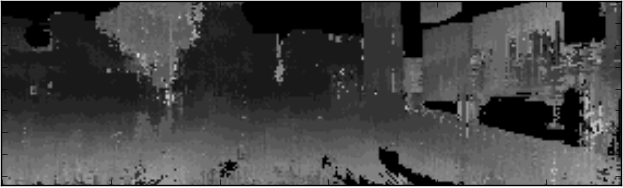}}\\
 \caption{Stereo image pair and estimated depth maps. Depth map scale ranges from 1 (Far, shown as black) to 25 (Near, shown as white).}
 \label{depth_map}
\end{figure}

A sample stereo image pair and the learned depth map is shown in Figure \ref{depth_map}.
In the figure, each predicted depth label is one pixel of the estimated depth map. 
An artifact of this depth estimation procedure is that 
object boundaries are expanded over their actual size due to the patch size used in the model. 
It can be also observed that the depth for feature-less regions like sky and plane surfaces 
is less accurate than in feature rich regions, because the model cannot detect any shift 
in those cases.

This is true, of course, for any disparity estimation scheme based on local region information,  
and when the goal is an explicit depth map, one should use a Markov Random Field or similar 
approach to cleaning up the obtained depth map. 
In the event where one is not interested in an exact depth map, but rather in depth cues 
to help make predictions that merely \emph{depend} on depth (similar to the bag-of-features
approach taken typically in motion estimation), 
a possible alternative is the use of an interest point detector as explained in Section~\ref{IPD}. 
Figure~\ref{masked} shows an example of an estimated depth map with interest points, and 
it shows that norm thresholding masks out most of the regions predominantly homogeneous 
regions in the image. 

In general, we thus observe that it is possible to infer depth information from the filter 
responses defined in Section~\ref{depth}, even if the information comes in the form of 
noisy cues, similar to most common estimates of motion, rather than in the form of a clean 
depth map.  We shall discuss an approach to exploiting this information in a bag-of-features 
pipeline for activity recognition in the next section. 

\subsection{Activity Recognition}

\begin{figure}
\centering
\begin{tabular}{c c}
 $W_q^X$ & \hspace{-10pt} \includegraphics[scale=1.45]{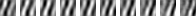} \vspace{3mm} \\
 $W_q^Y$ & \hspace{-10pt} \includegraphics[scale=1.45]{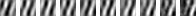} \\
\end{tabular}
\caption{Example of a filter pair learned on sequences by the SAE-D model from the Hollywood3D dataset.} 
\label{filterSAE}
\end{figure}

We evaluate the effect of implicit depth encoding on the task of activity recognition, using 
the Hollywood3D dataset introduced by \cite{Hadfield13}. The dataset consists of 
stereo video sequences along with computed depth videos. The videos are of
$14$ different categories with $643$ videos for training and $302$ for testing. The different categories are 
'Run', 'Punch', 'Kick', 'Shoot', 'Eat', 'Drive', 'UsePhone', 'Kiss', 'Hug', 'StandUp', 'SitDown', 'Swim', 'Dance' and 'NoAction'.
The videos are downsampled spatially from size of $1920 \times 1080$ to $320 \times 240$. Models are trained on 
PCA whitened spatio-temporal block pairs with each block of size $10\times 16 \times 16$. 
$300,000$ samples are used for training and the number of hidden units is fixed for all models to $300$.
A sample feature pair learned by the SAE-D model is shown 
in Figure \ref{filterSAE}. Each filter in the pair spans ten frames. 
The filters are again Gabor-like and show a continuous phase shift through time, 
and another phase shift across camera views. 

\begin{figure}
 \centering
\hspace{-20pt} \includegraphics[scale=0.71]{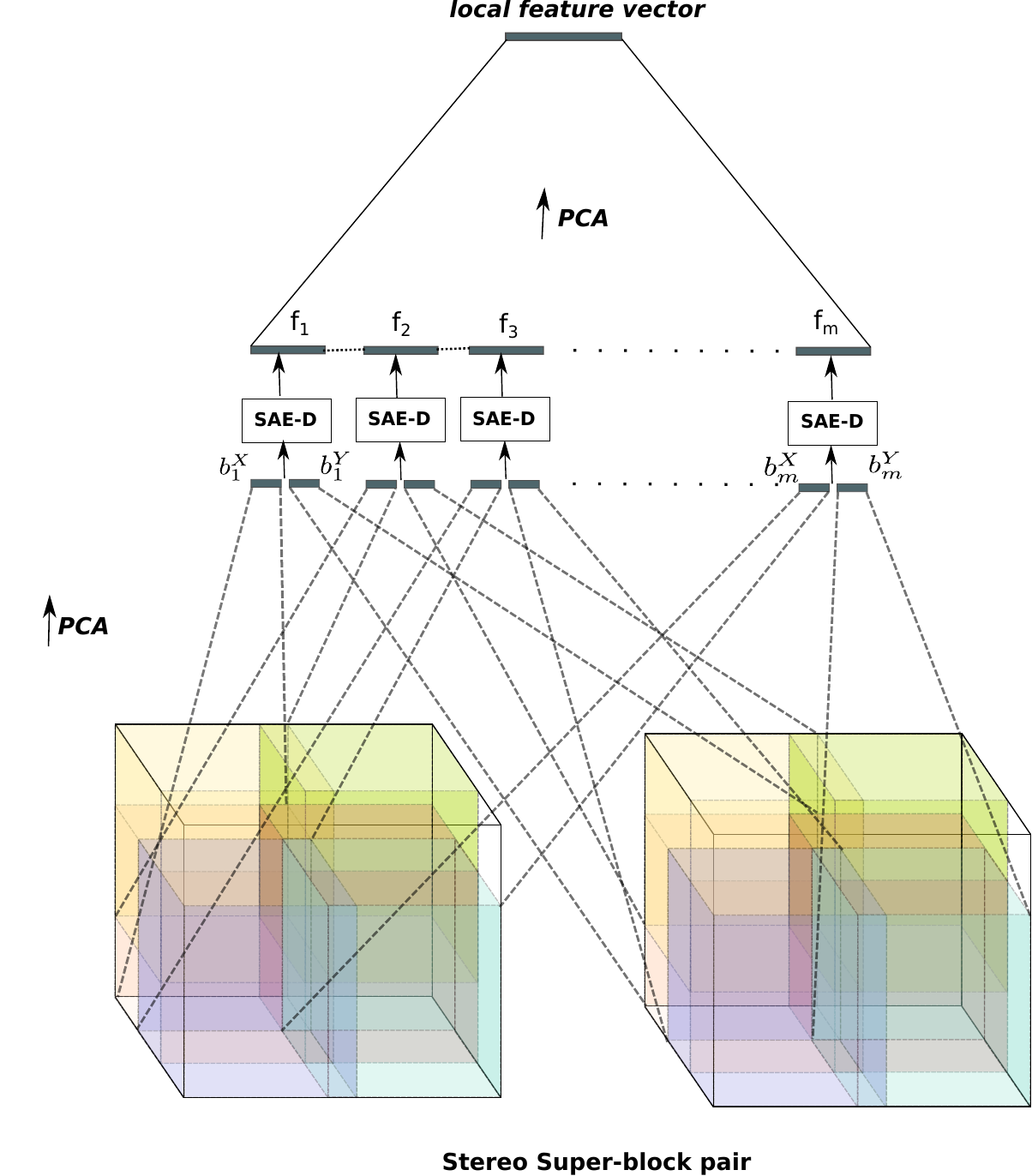}
 \caption{Feature extraction framework using the SAE-D model.}
 \label{inference}
\end{figure}

\begin{figure*}
 \centering
 \includegraphics[scale=0.75]{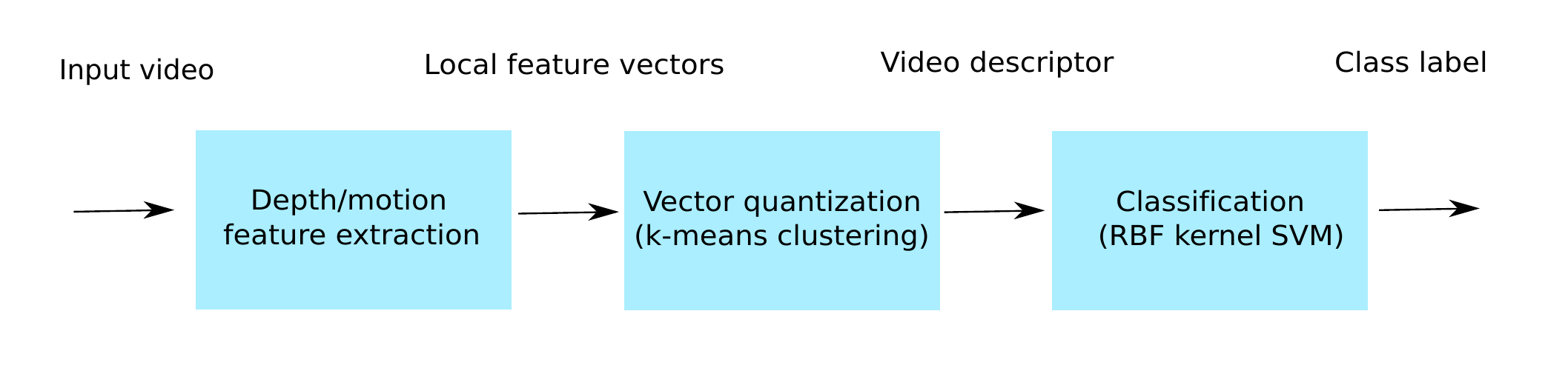}
 \caption{Flow diagram of the classification pipeline.}
 \label{pipeline}
\end{figure*}

For the quantitative evaluation, we use the framework presented by \cite{Hadfield13}.  
After performing feature extraction, we perform K-means vector quantization 
followed by a multi-class SVM with RBF kernel for classification. 
A flow diagram of the pipeline is visualized in Figure \ref{pipeline}.

Feature are extracted using a convolutional architecture similar to that 
presented in \cite{synchrony, ISA}. Super block pairs of size
$14\times 20 \times 20$ pixels each are cropped densely with stride $(7,10)$ in time and 
space, respectively, from the stereo video pairs.
From the super blocks, sub-blocks of the same size as the training block 
size ($10 \times16 \times16$ pixels) are cropped with stride $(4,4)$, resulting 
in $8$ sub-blocks per super block.

We first compute the feature vector for each stereo sub-block pair.  
We then concatenate feature vectors corresponding to the sub-blocks of a super block
and reduce their dimensionality using PCA. 
This procedure, using the SAE-D model as an example, is visualized in 
Figure \ref{inference}. The number of words for K-means vector quantization 
is set to $3000$. 

Our main goal in these experiments is to evaluate the impact of the implicit 
depth encoding in the task of activity recognition.
We compare a variety of settings to this end. 
In {\bf experiment 1}, the SAE-D is used for feature extraction. As we discussed 
in Section~\ref{depth} the SAE-D primarily encodes depth.
{\bf Experiment 2} uses the SAE-M for feature extraction with only one of the stereo 
channels as input. 
{\bf Experiment 3} employs the SAE-MD for features extraction, and is thus based on a 
representation that integrates across-frame and across-channel correlations. 
In {\bf Experiment 4} we test two alternative ways of integrating depth and motion information, 
by combining the representations from two separately trained SAE-D and SAE-M models. 
The first, which we call SAE-MD(Ct), amounts to concatenating the representations 
from SAE-D and SAE-M as features. 
The other, SAE-MD(Av), amounts to computing the average precision using the mean 
over confidences from experiments 1 and 2. Thus it amounts to averaging the 
classification decisions of two separate classification pipelines (one based primarily 
on depth, and the other based primarily on motion). 

Each configuration is evaluated by computing the average precision and the 
correct-classification rates. 
The results are reported in Tables \ref{AP} and \ref{CCRate}.
We repeated the experiments using the norm-thresholding interest point detector 
described in Section \ref{IPD}.

From the results it can be observed that the combination of motion and depth 
cues performs better than using individual cues. 
All results, including the motion-only, the depth-only and the combination models, 
outperform all existing models, based on hand-crafted representations, 
by a very large margin, and they are to the best of our knowledge the best 
reported results on this task to date. 

It has been observed in the past that learning based features tend to outperform 
more traditional features, like SIFT, in object recognition tasks and, more 
recently and by a larger margin, in motion analysis tasks  
as compared to spatio-temporal variations 
of SIFT (e.g., \cite{ISA,Taylor:2010,synchrony}).
This observation is confirmed in this 4-D dataset, where it seems to be even more pronounced. 

We can also observe that models using interest points (cf., Section~\ref{IPD}) 
provide an additional consistent (albeit smaller) improvement over those that do not. 
Furthermore, the use of depth information provides an edge over motion-only models. 
Interestingly, the overall effect of the various variations of the model differ 
heavily across action class, which can be seen in Table~\ref{AP}.  
For example, the AP for classes Run, Kick, Shoot and Eat are the highest when using 
primarily depth features for classification; NoAction and Kiss have best AP when using 
just motion features; 
and the AP for all the other classes is the highest when combining depth and motion features. 
This can be due to multiple reasons, and it is likely related to the average 
depth variation within the activity class. 
A detailed analysis of which type of information is the most useful for which 
type of activity class is an interesting direction for future work. 

A well-known and popular ``recipe'' to improve performance in learning tasks has been 
to base classification decisions on the combination of multiple different models, each 
of which utilizes a different type of feature. While this recipe often works well in
practice, the main challenge to make it work is to develop models which are 
sufficiently \emph{different} from one another, so they yield a sufficiently large 
reduction of variance. Utilizing the combination of depth and motion cues may be viewed 
in this context also as a way to extract cues from video data which are different, 
since they represent very different properties of the environment. 

\begin{table*}
\begin{center}
\small
\begin{tabular}{|l||c|c|c|c|c|||c|c|c|}
 \hline
 \bf{Action} 	& \bf{SAE-MD} & \bf{SAE-MD(Av)} & \bf{SAE-MD(Ct)} & \bf{SAE-M} & \bf{SAE-D} & \bf{3D-Ha} & \bf{4D-Ha} & \bf{3.5D-Ha}\\ \hline\hline
 \it{NoAction}	& 12.10 & 12.77 & 13.10 & \bf{15.73} & 12.15 & 12.1 & 12.9 & 13.7\\ 
 \it{Run}	& 52.56 & 50.44 & 51.45 & 45.38 & \bf{56.07} & 19.0 & 22.4 & 27.0\\ 
 \it{Punch}	& \bf{41.09} & 38.01 & 32.68 & 33.86 & 36.17 & 10.4  &  4.8 &  5.7\\ 
 \it{Kick}	&  9.41 &  7.94 &  6.86 &  6.63 & \bf{11.84} &  9.3  &  4.3 &  4.8\\ 
 \it{Shoot}	& 30.26 & 35.51 & 30.49 & 30.52 & \bf{40.72} & 27.9  & 17.2 & 16.6\\  
 \it{Eat}	&  5.85 &  7.03 &  6.78 &  7.29 &  \bf{9.03} &  5.0  &  5.3 &  5.6\\ 
 \it{Drive}	& 52.65 & 59.62 & 51.35 & 61.61 & 45.19 & 24.8  & 69.3 & \bf{69.6}\\ 
 \it{UsePhone}	& 22.79 & \bf{23.92} & 19.01 & 23.60 & 23.36 &  6.8  &  8.0 &  7.6\\ 
 \it{Kiss}	& 15.03 & 16.40 & 16.12 & \bf{17.86} & 17.06 &  8.4  & 10.0 & 10.2\\ 
 \it{Hug}	&  6.64 &  7.02 &  7.61 &  7.38 &  9.27 &  4.3  &  4.4 & \bf{12.1}\\ 
 \it{StandUp}	& \bf{37.35} & 34.23 & 37.01 & 29.16 & 15.01 & 10.1  &  7.6 &  9.0\\ 
 \it{SitDown}   &  6.51 &  6.95 &  7.53 & 7.40 &  \bf{9.06} &  5.3  &  4.2 &  5.6\\ 
 \it{Swim}	& 16.58 & \bf{29.48} & 17.60 & 29.45 & 26.70 & 11.3  &  5.5 &  7.5\\ 
 \it{Dance}	& 43.15 & 36.26 & \bf{44.59} & 29.64 & 25.12 & 10.1  & 10.5 &  7.5\\ \hline
 \bf{mean AP}   & 25.14 & \bf{26.11} & 24.45 & 24.61 & 24.05 & 12.6  & 13.3 & 14.1\\ \hline
\end{tabular}
\end{center}
\caption{Average precision per class on the Hollywood 3D action dataset. 
The APs for the bag of words descriptor using the 3D-Ha, 
4D-Ha, 3.5D-Ha interest points are reported from \cite{Hadfield13}. 
The values in bold are the best AP per class across all methods.}
\label{AP}
\end{table*}

\begin{table}
\begin{center}
\small
\begin{tabular}{|l|c|c|c|}
 \hline
 \bf{Method} 	& \bf{Interest points} & \bf{AP} & \bf{CC Rate}\\ \hline\hline
 SAE-M		& None & 24.31 & 29.61\\ \hline
 SAE-MD		& None & 24.69 & 29.47\\ \hline
 SAE-D		& None & 23.53 & 26.82\\ \hline
 SAE-M		& N-Th & 24.61 & 31.79\\ \hline
 SAE-MD		& N-Th & 25.14 & 30.46\\ \hline
 SAE-D		& N-Th & 24.05 & 26.49\\ \hline
 SAE-MD(Ct)	& N-Th & 24.45 & 29.47\\ \hline
 SAE-MD(Av)	& N-Th & 26.11 & 30.13\\ \hline
 HoG/Hof/HoDG \cite{Hadfield13}	& 3.5D-Ha & 14.1 & 21.8 \\ \hline
 RMD-4D\cite{Hadfield13}	& 3D-Ha & 15.0 & 15.9 \\ \hline
 \end{tabular}
\end{center}
\caption{Correct classification rate and average precision.}
\label{CCRate}
\end{table}

\section{Discussion}
\label{discussion}
Most current practical work on stereopsis focuses on extracting dense depth-maps using MRFs.
Potential reasons for biology to take a different route might be that (a) depth via deep learning makes
it possible to use the exact same learning algorithm for depth inference that is 
also used to recognize objects and motion; 
(b) a simple depth cue, as given by a feature vector, $H$, is often entirely sufficient 
to take swift vital decisions, such as to dodge an approaching object; 
(c) learning depth inference from data allows for feed-forward depth perception, 
and thus to avoid the need for a complicated and brittle pipeline, which involves 
rectification, hypothesis generation, and robustification using RANSAC \cite{Hartley2004}. 

In this paper we showed how unsupervised feature learning may be used to mimic this way of
extracting depth cues from image pairs, and that  
learning joint representations of motion and depth within a single type of architecture 
and a single type of learning rule can achieve state-of-the-art performance 
in a 3-D activity recognition task. 
Our work is to the best of our knowledge the first published work that shows that 
deep learning approaches, which have hitherto been shown to work well in object 
and motion recognition tasks, are also applicable in the domain of depth inference, 
or more generally to 3-D vision. 

{\small
\bibliographystyle{ieee}
\bibliography{ref}

\begin{thebibliography}{10}\itemsep=-1pt

\bibitem{Adelson85}
E.~H. Adelson and J.~R. Bergen.
\newblock Spatiotemporal energy models for the perception of motion.
\newblock {\em J. OPT. SOC. AM. A}, 2(2):284--299, 1985.

\bibitem{OlshausesCadieu}
C.~F. Cadieu and B.~A. Olshausen.
\newblock {Learning Intermediate-Level Representations of Form and Motion from
  Natural Movies}.
\newblock {\em Neural Computation}, 24(4):827--866, Dec. 2011.

\bibitem{FleetBinocular}
D.~Fleet, H.~Wagner, and D.~Heeger.
\newblock {Neural encoding of binocular disparity: Energy models, position
  shifts and phase shifts}.
\newblock {\em Vision Research}, 36(12):1839--1857, June 1996.

\bibitem{fleet1991phase}
D.~J. Fleet, A.~D. Jepson, and M.~R. Jenkin.
\newblock Phase-based disparity measurement.
\newblock {\em CVGIP: Image understanding}, 53(2):198--210, 1991.

\bibitem{Geiger2012CVPR}
A.~Geiger, P.~Lenz, and R.~Urtasun.
\newblock Are we ready for autonomous driving? the kitti vision benchmark
  suite.
\newblock In {\em Conference on Computer Vision and Pattern Recognition
  (CVPR)}, 2012.

\bibitem{Hadfield13}
S.~Hadfield and R.~Bowden.
\newblock Hollywood 3d: Recognizing actions in 3d natural scenes.
\newblock In {\em Proceeedings, conference on Computer Vision and Pattern
  Recognition}, Portland, Oregon, June23 - 28 2013.

\bibitem{Hartley2004}
R.~I. Hartley and A.~Zisserman.
\newblock {\em Multiple View Geometry in Computer Vision}.
\newblock Cambridge University Press, ISBN: 0521540518, second edition, 2004.

\bibitem{5539947}
C.~Kanan and G.~Cottrell.
\newblock Robust classification of objects, faces, and flowers using natural
  image statistics.
\newblock In {\em Computer Vision and Pattern Recognition (CVPR), 2010 IEEE
  Conference on}, pages 2472--2479, 2010.

\bibitem{synchrony}
K.~R. Konda, R.~Memisevic, and V.~Michalski.
\newblock The role of spatio-temporal synchrony in the encoding of motion.
\newblock {\em CoRR}, abs/1306.3162, 2013.

\bibitem{ISA}
Q.~Le, W.~Zou, S.~Yeung, and A.~Ng.
\newblock Learning hierarchical invariant spatio-temporal features for action
  recognition with independent subspace analysis.
\newblock In {\em CVPR}, 2011.

\bibitem{gated}
R.~Memisevic.
\newblock Gradient-based learning of higher-order image features.
\newblock In {\em ICCV}, 2011.

\bibitem{multiview}
R.~Memisevic.
\newblock On multi-view feature learning.
\newblock In {\em Proceedings of the 29th International Conference on Machine
  Learning (ICML-12)}, pages 161--168, 2012.

\bibitem{ohzawa1990stereoscopic}
I.~Ohzawa, G.~C. Deangelis, and R.~D. Freeman.
\newblock Stereoscopic depth discrimination in the visual cortex: neurons
  ideally suited as disparity detectors.
\newblock {\em Science}, 249(4972):1037--1041, 1990.

\bibitem{qian1994computing}
N.~Qian.
\newblock Computing stereo disparity and motion with known binocular cell
  properties.
\newblock {\em Neural Computation}, 6(3):390--404, 1994.

\bibitem{contractiveAE}
S.~Rifai, P.~Vincent, X.~Muller, X.~Glorot, and Y.~Bengio.
\newblock {Contractive Auto-Encoders: Explicit Invariance During Feature
  Extraction}.
\newblock In {\em ICML}, 2011.

\bibitem{scharstein2002taxonomy}
D.~Scharstein and R.~Szeliski.
\newblock A taxonomy and evaluation of dense two-frame stereo correspondence
  algorithms.
\newblock {\em International journal of computer vision}, 47(1-3):7--42, 2002.

\bibitem{Taylor:2010}
G.~W. Taylor, R.~Fergus, Y.~LeCun, and C.~Bregler.
\newblock Convolutional learning of spatio-temporal features.
\newblock In {\em Proceedings of the 11th European conference on Computer
  vision: Part VI}, ECCV'10, 2010.

\end{thebibliography}
}

\end{document}